\documentclass[a4paper, 10pt, conference]{ieeeconf}      
\usepackage{FG2025}
\usepackage{graphicx}
\usepackage{url} 
\usepackage{multirow}
\usepackage{booktabs}
\usepackage{subfigure}

\FGfinalcopy % *** Uncomment this line for the final submissio
\IEEEoverridecommandlockouts                              
\def\FGPaperID{136} % *** Enter the FG2025 Paper ID here

\title{\LARGE \bf
iLSU-T: an Open Dataset for\\
Uruguayan Sign Language Translation 
}

\author{\parbox{16cm}{\centering
    {\large Ariel E. Stassi$^{1,*}$, Yanina Boria$^{1,2}$, J. Matías Di Martino$^{3,4,+}$, and Gregory Randall$^{1,+}$}\\
    {\normalsize
    $^1$Universidad de la Rep\'{u}blica, Uruguay.
    $^2$Universidad de Buenos Aires, Argentina.\\
    $^3$Universidad Cat\'{o}lica del Uruguay.
    $^4$Duke University, USA.\\
    $^*$Corresponding author: \url{astassi@cup.edu.uy}
    $^+$Co-senior authors.
}}
}

\usepackage{fancyhdr}
\begin{document}

\ifFGfinal
\thispagestyle{empty}
\pagestyle{empty}
\else
\author{Anonymous FG2025 submission\\ Paper ID \FGPaperID \\}
\pagestyle{plain}
\fi
\maketitle

% AGREGADO CAMERA READY
\thispagestyle{fancy}

%%%%%%%%%%%%%%%%%%%%%%%%%%%%%%%%%%%%%%%%%%%%%%%%%%%%%%%%%%%%%%%%%%%%%%%%%%%%%%%%
\begin{abstract}
Automatic sign language translation has gained particular interest in the computer vision and computational linguistics communities in recent years. 
Given each sign language country's particularities, machine translation requires local data to develop new techniques and adapt existing ones. This work presents iLSU-T, an open dataset of interpreted Uruguayan Sign Language RGB videos with audio and text transcriptions. This type of multimodal and curated data is paramount for developing novel approaches to understand or generate tools for sign language processing. iLSU-T comprises more than 185 hours of interpreted sign language videos from public TV broadcasting. It covers diverse topics and includes the participation of 18 professional interpreters of sign language.  
A series of experiments using three state-of-the-art translation algorithms is presented. The aim is to establish a baseline for this dataset and evaluate its usefulness and the proposed pipeline for data processing. The experiments highlight the need for more localized datasets for sign language translation and understanding, which are critical for developing novel tools to improve accessibility and inclusion of all individuals. Our data and code can be accessed at \mbox{\url{https://github.com/ariel-e-stassi/iLSU-T}}. 
\end{abstract}

%%%%%%%%%%%%%%%%%%%%%%%%%%%%%%%%%%%%%%%%%%%%%%%%%%%%%%%%%%%%%%%%%%%%%%%%%%%%%%%%
\section{INTRODUCTION}
\label{sec:intro}
\noindent % es necesario?
Sign languages are natural languages of the %that occur in 
deaf communities worldwide that use manual and non-manual features over time and 3D space to convey meaning. Manual features include hand shapes, locations, orientations, and movements. Non-manual features include facial expressions, lip patterns, gaze, and body movements. People without hearing impairment generally do not know sign language, so automatic translation can shorten the communication gap between signers and listeners. Moreover, it can lower the cost of the automatic translation and generation of media content that includes sign language~\cite{bragg_sign_2019, de_coster_machine_2023}. 
% In the last years sign language processing has gained special interest in the scientific community of computer vision and natural language processing. Sign language processing can be defined as an umbrella term that includes sign language recognition, translation, and production~\cite{bragg_sign_2019}.

Each region or country has its sign language, its lexicon, grammar rules, and dialect. LSU (an acronym for \textit{Lengua de Señas Uruguaya}) is the sign language used by the deaf community in Uruguay. Suitable data is required to develop solutions for LSU processing tasks, including LSU automatic translation.

In this paper, we present the first dataset for automatic processing of LSU, with particular interest in tackling the problem of automatic translation of interpreted RGB videos using different data sources. %As we discuss later, it is not the same thing to translate to a written language based on natural signing data than interpreted sign language one. In the later case, the interpreter introduce some variations w.r.t. the original spoken message, among others the space and time limitations, variations in the comprehension of the interpreters, etc.
The main contributions of this work are: 
\begin{itemize}
\item iLSU-T, the first dataset with multimodal video, audio, and text for LSU translation. iLSU-T comprises more than 185 hours of curated video from TV broadcasting in Uruguay. 
    
\item A preprocessing pipeline to derive the \mbox{iLSU-T} dataset.

\item A theoretical discussion from the linguistic perspective about the problem of aligning and annotating interpreted sign language videos with text.

\item The first recorded evaluation and benchmarking of state-of-the-art available methods for sign language translation in the LSU context.  
\end{itemize}

\section{RELATED WORK}
\label{sec::related_work}
\noindent
Sign language processing is a set of techniques for analysis and understanding sign language data, including recognition, translation, and sign language production \cite{bragg_sign_2019, de_coster_machine_2023}. Sign language processing is a naturally interdisciplinary field that lies at the intersection between computer vision, machine translation, and linguistics \cite{de_coster_machine_2023}. Among the problems associated with sign language processing, we can mention sign language (or fingerspelling) detection, i.e., recognizing whether a signer appears in a video doing sign language~\cite{moryossef2020real} (or fingerspelling~\cite{shi2021fingerspelling}, respectively). On the other hand, there is the problem of recognizing signs, either isolated or within a sequence. More specifically, the problems of isolated sign language recognition~\cite{de-coster-etal-2020-sign, De_Coster_2021_CVPR, jiang2021skeleton}, continuous sign language recognition{~\cite{zhu2024chinese}}, and sign spotting~\cite{wong2022hierarchical}. In the case of a continuous stream of sign language content, several existing techniques require pre-segmentation of the data into phrases or signs depending on the downstream task~\cite{camgoz2018neural, camgoz2020sign, voskou2021stochastic, yin2023gloss}. The automatic approach to tackle this problem has been named sign language segmentation~\cite{bartoli_automatic_2020,moryossef-etal-2023-linguistically,Renz2021signsegmentation_a}. 
Continuous sign language recognition recognizes the gloss sequence in the input sign language phrases. In this case, the labels are gloss annotations, defined as a written representation of sign language content based on the chronologically labeled sign language units in a one-to-one fashion~\cite{yin2023gloss}. 

\begin{table*}[t!]
\caption{Sign Language Datasets for sign language translation (sorted by number of hours). \\Number of signers, hours, samples, and vocabulary size (used words).}
\label{table_example}
\centering
\resizebox{0.97\textwidth}{!}
{\begin{tabular}{lccccccccccc}

\toprule

{Dataset}             &   {Source language}    &   Target language   & \#signers    &  \#hours &  \#samples &  Vocabulary &  Video quality   &   Annotations  &   Source   \\
\midrule

Phoenix2014T~\cite{camgoz2018neural} &    DGS           &       German        &  9           & 10.5              &     8257   &  2k9 & 210$\times$260@25\,fps   &    text, gloss &    TV \\
LSA-T~\cite{dal2022lsa}              &    LSA     &       Spanish       & 103          & 21.8             &    14880   & 14k2 &  1920$\times$1080@30\,fps &     text (SD)  & Web \\
CSL-Daily~\cite{zhou2021improving} & CSL & Chinese & 10 & 23 & 20654 & 2k5 & 1920$\times$1080@30\,fps & text, gloss & Lab\\
KETI~\cite{ko2019neural}             &        KSL       &       Korean        &  14          & 28                &    14672 &  419 &  1920$\times$1080@30\,fps &    text        & Lab\\
AUSLAN-Daily~\cite{shen2024auslan}   &     Auslan                &  English                   &    67          &     45              &     25106    &  13k9 &       1280$\times$720/1920$\times$1080@25\textbar30\,fps         &      text          &   TV \\        
SIGNUM{~\cite{koller2015continuous}}                    &        DGS       &        German       &  25          &  55.3             &    33210   & N/A &  776$\times$578@30\,fps    &    text        & Lab\\
How2Sign~\cite{Duarte_CVPR2021}      &     English                   &           ASL          &    11          &        79           &    35k2   &  16k   &    1280$\times$720@30\,fps                  &    text          &    Lab \\        

% CSL100~\cite{huang2018video}         &        CSL      & Chinese             &    50        &  100+             & 25000  &  178  & 1920$\times$1080@30\,fps  &   text         & Indoor  \\
OpenASL~\cite{shi2022open}           &   ASL          &    English          &  220  &   288             &  98417     & 33k5 & variable &   text          & Web \\

BOBSL~\cite{albanie2021bbc}          &   English              & BSL          &  37          &   1467           &  1M2     & 78k   & 444$\times$444@25\,fps   &   text          &  TV \\
\midrule
\textbf{iLSU-T (ours) }                       &  \textbf{Spanish}               &            \textbf{LSU}      &   \textbf{18}         &   \textbf{201.5}           & \textbf{86550}      & \textbf{37k9} & \textbf{variable, 343$\times$364@25\textbar30\,fps} &    \textbf{text (SD)}    &  \textbf{TV}\\
\bottomrule
\end{tabular}}
\label{table:slt_datasets}
\end{table*}

Sign language translation (SLT) maps a sequence of signs in a sentence to the corresponding written phrase, including the target language's grammar. SLT methods can be coarsely classified into three categories~\cite{de_coster_machine_2023, yin2023gloss}: 1) two-stage methods based on continuous sign language recognition followed by a gloss-to-text translation; 2) end-to-end gloss supervised methods; and 3) end-to-end gloss-free methods. Gloss annotations help the models to learn the alignment between signs and (visual) input features, but their generation requires significant expert annotation efforts. Hence, gloss-based approaches are frequently limited in the coverage of different domains, making it challenging to apply them in realistic scenarios~\cite{lin-etal-2023-gloss}. In this work, we are focused on gloss-free sign language data and, hence, translation methods. 

Table~\ref{table:slt_datasets} shows the most popular and recent datasets for sign language translation sorted by size. The acronyms used in the source and target language columns refer to the local sign language. Most table datasets were constructed using original sign language data with audio or subtitles. The iLSU-T dataset is the first large-scale dataset for automatic LSU translation. The table shows that it is comparable to other large state-of-the-art datasets regarding the number of samples, duration, vocabulary size, and signers. Note that SD in the annotations column refers to ``subtitle derived'' with particularities in the phrase conformation.

%%%%%%%%%%%%%%%%%%%%%%%%%%%%%%%%%%%%%%%%%%%%%%%%%%%%%%%%%%%%%%%%%%%%%%%%%%%%%%%%
\section{iLSU-T dataset}

\subsection{Data sources}
\noindent
The data sources of iLSU-T videos are two channels of the public Uruguayan Television --Canal~5 and TV~Ciudad, hereafter referred to as Sources 1 and 2, respectively-- and sessions of the Uruguayan Parliament --hereafter referred to as Source 3--, with an average width and height of 343$\times$364 pixels, respectively %, and two values of frame rate, 25 and 30~fps~
(see Section~\ref{subsec:dataset_statistics} for more details). 

\subsection{Processing pipeline and data curation}
\label{subsec:signlanguageprocessing}
\begin{figure*}
    \centering
    \subfigure[]{\includegraphics[trim=4.5cm 1.5cm 11cm 0cm, clip=true, height=0.285\linewidth]{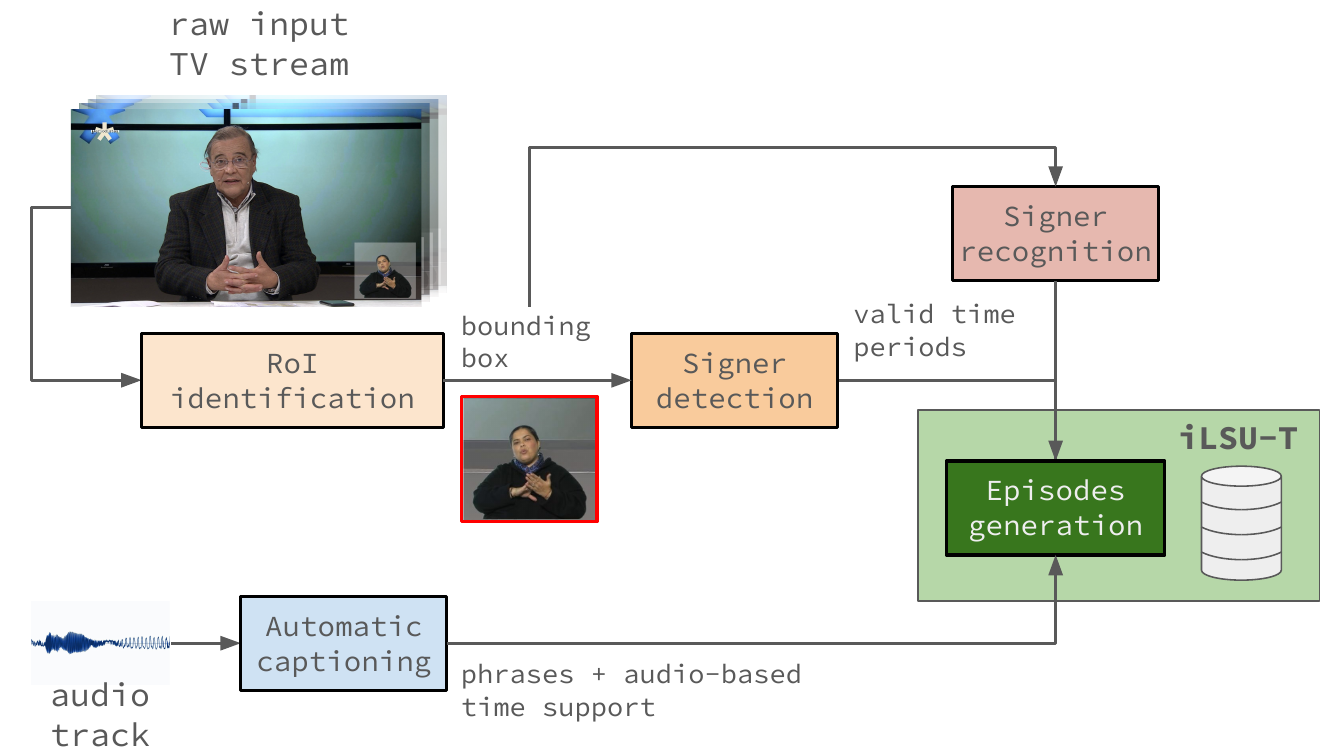}}
    \subfigure[]{\includegraphics[trim=9cm 1.5cm 8cm 0cm, clip=true, height=0.285\linewidth]{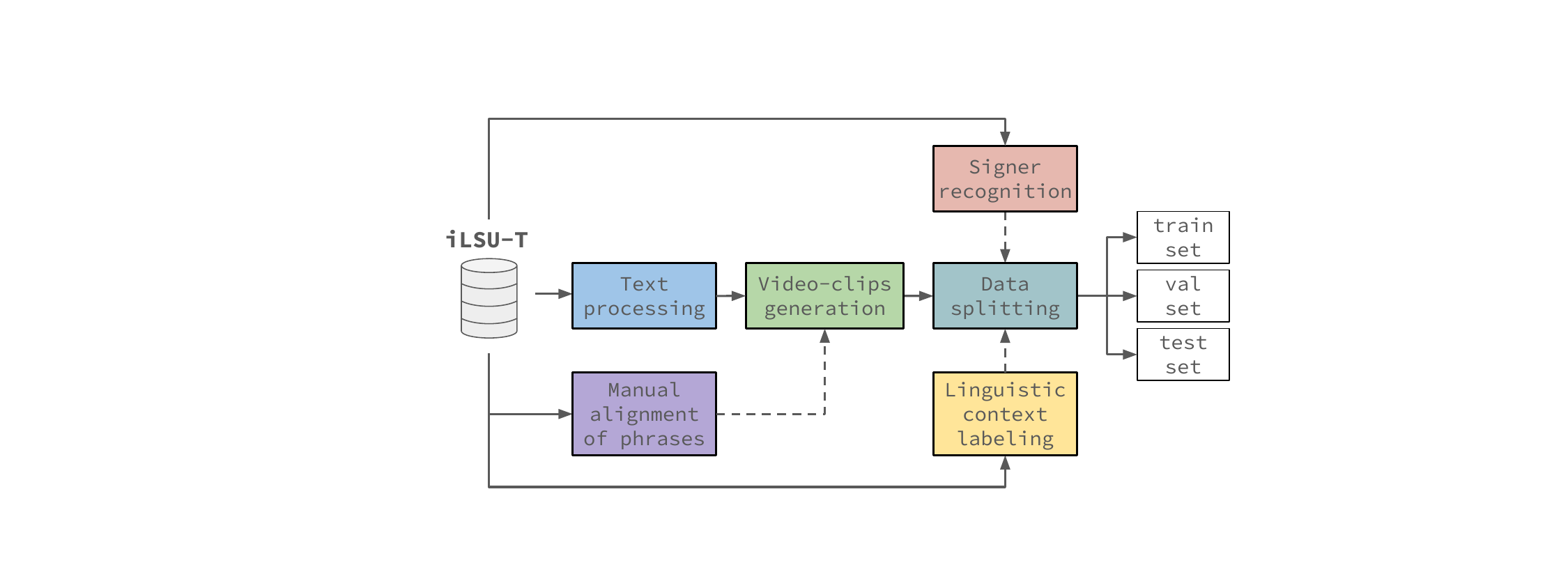}}

    % \begin{subfigure}[t]{0.49\textwidth}
    %     \centering
    %     \includegraphics[trim=4.5cm 1.5cm 11cm 0cm, clip=true, height=0.58\linewidth]{iLSU_T_pipeline-a.pdf}
    %     \caption{iLSU-T dataset, episodes generation.}
    % \end{subfigure}
    % \begin{subfigure}[t]{0.49\textwidth}
    %     \centering
    %     \includegraphics[trim=9cm 1.5cm 8cm 0cm, clip=true, height=0.58\linewidth]{iLSU_T_pipeline-b.pdf} 
    %     \caption{iLSU-T dataset, possible uses for translation.}
    % \end{subfigure}       
    
    \caption{(a) Pipeline for the generation of iLSU-T dataset. First, we locate the sign language interpreter Region of Interest (RoI) in the raw videos. Then, the presence and recognition of the signer are determined. From the audio track, text phrases are obtained by automatic captioning (transcription). (b) An experimental pipeline was implemented to benchmark iLSU-T for automatic translation. The text processing step refers to changes in the time support of text events to compose aligned phrases. Dashed lines denote auxiliary stages to carry out controlled experiments following different criteria, for example, by splitting the data considering the signer's IDs or by restrictions in the linguistic diversity from the context labeling.}
    \label{fig:preprocessing_pipeline}
\end{figure*}

\noindent
We define an episode as a single video containing a continuous broadcast block (in a similar way as~\cite{albanie2021bbc}). Here, we imposed that each episode be signed by only one interpreter, i.e., its temporal boundaries be fixed based on the signer's appearance on the scene or when there is a signer substitution.
Given the raw data, we use a processing pipeline (see Fig.~\ref{fig:preprocessing_pipeline}) to compose valuable episodes. %Fig.~\ref{fig:preprocessing_pipeline} shows the pipeline steps. 
The data generation process includes five main stages: (1)~RoI identification, (2) Signer recognition, (3) Automatic captioning, (4) Manual alignment of phrases, and (5) Linguistic context labeling.

\subsubsection{RoI identification}
\begin{figure}[b!]
    \centering
    \includegraphics[trim=1cm 1cm 7cm 1cm, clip=true,width=0.9\linewidth]{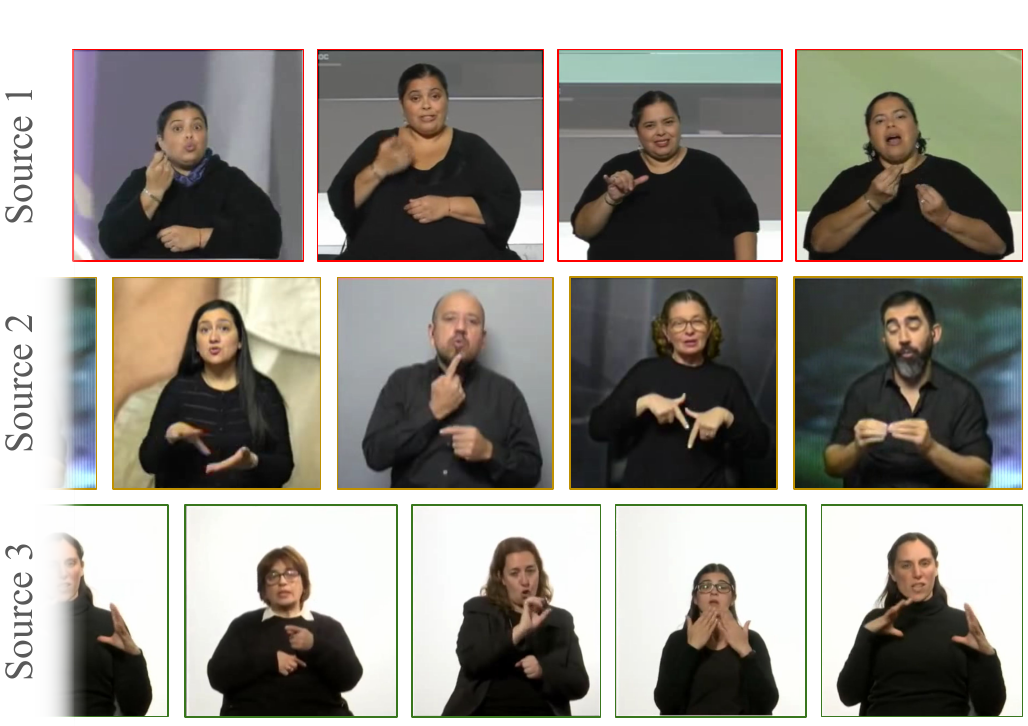}
    \caption{Isolated frame examples of RoI, background, and signers for the three video sources of iLSU-T.}
    \label{fig:datasources}
\end{figure}

The RoI (Region of Interest) corresponds to the sign language interpreter's bounding box in each video. As each raw file has only one RoI position for the entire video, it was manually labeled by visual inspection. We use the $(x,y)$ coordinates of the rectangle's upper left and lower right corners. Each RoI includes only one signer. Fig.~\ref{fig:datasources} shows RoI examples at each source's frame level. Please note the different RoI backgrounds depending on the media source. Note that the sizes and aspect ratios are variable between the sources and even between different episodes from the same source. Additionally, the interpreter scale within the RoI presents slight variations across episodes.

\subsubsection{Signer recognition} To detect and recognize if a given signer is present in the RoI, we used a KNN-based face classifier\footnote{\url{https://github.com/ageitgey/face_recognition/blob/master/examples/face_recognition_knn.pys}.}. The classifier was trained in a supervised manner with 50 samples per signer. All the videos were processed by a uniform sampling of one frame per second, and the corresponding signer was classified. A median filter was applied for post-processing recognition, considering that the minimum time per signer was 30 seconds. The raw videos were segmented to have one signer per episode. 
Finally, the signer recognition stage was verified, and time boundaries were refined for each episode by visual inspection. 

\subsubsection{Automatic captioning} Text subtitles were produced using WhisperX~\cite{bain2022whisperx} over the audio track, using the \texttt{large-v3} model, which provides text segmented in sentences with timestamps at the word level. 
% The audio captioning was done twice, one by considering the audio track corresponding to the video track as is and the other by considering a 10-second negative offset at the beginning of each video. The 10--seconds value was selected following a linguistic study about the effects of the lag time in the committed errors by the interpreters who reported a maximum onset lag time of 6 seconds~\cite{cokely1986effects}. The 10-second delayed track is particularly useful for obtaining the text and context in the initial periods of each episode. %Note that, by design, each episode begins when the signer appears on the scene or when there is a signer substitution.

\subsubsection{Manual alignment of phrases}
\noindent
Simultaneous interpretation encompasses interpretation %between languages of different modalities, 
from an audio-oral language to a viso-gestural language~\cite{de2015traduccao}. The resulting linguistic form exhibits specific characteristics that must be considered to determine the appropriate alignment between text and sign language gestures. Naturally, the interpretation of LSU will inevitably be out of sync with the transcribed text. Moreover, there is a variable delay between both modalities. 

As presented in Section~\ref{sec:intro}, SLT maps a sign language phrase to the corresponding written phrase. Then, the segmentation of the video and text content and its mutual alignment, considering the video and the audio or text tracks, are required. The alignment task can be defined as a comparative translation task involving the analysis of the expressions in LSU and Spanish and matching them. The segmentation task must determine where the LSU content can be cut into phrases or utterances. Text is automatically segmented based on recognizing speech pauses. However, this segmentation is not necessarily aligned with the pauses used in the sign language interpretation. In this work, we segmented LSU content based on two concepts: pauses and epenthesis. 

Pauses in interpreting may be attributed to several factors: the discourse itself, instances of overlapping speech, the time required for the interpreter to process the spoken information, technical difficulties, and other variables. These pauses are expressed in a variety of ways: (1)~the interpreter remains stationary after a given sign and then resumes interpreting from that point, or (2)~the interpreter returns to the \textit{resting position}\footnote{In this work the \textit{resting position} is conceived in the same way as in~\cite{stassi2022lsu}.}. The involved frames do not correspond to linguistic segments \textit{per se}; instead, they represent strategies employed in simultaneous interpreting from spoken languages into sign languages. 

In the literature on sign language phonology, the term \textit{epenthesis} is associated with the phonological feature ``movement'', specifically concerning interpolation transitions that occur between two signs made at two different places of articulation~\cite{geraci2009epenthesis, liddell1984think}. These movements differ systematically from those encoded in sign language phonology~\cite{brentari1998prosodic}. Furthermore, epenthetic movements can be made from and to the resting position. In both cases, this phenomenon allows for segmenting sign language content in phrases or utterances without affecting their meaning. 

%Considering the discussed characteristics, 
A sign language translation human expert carried out the manual alignment of phrases in the following way: (1) Phrase beginnings and endings are removed. The resting position and the initial and final epenthetic movements of each segment in LSU are eliminated. (2) Textual segments not interpreted in LSU are eliminated. (3) Consider an instance where two text segments are present in the source text. Still, only a single segment exists in LSU characterized by a sustained signer activity, i.e., no pauses. Then, the video segments are separated by the epenthetic movement between the two LSU phrases. (4) If the interpreter pauses in the middle of a clip that cannot be cut, and in this pause assumes the resting position, a ``0'' mark is made to consider this annotation in future dataset use. %experiments design. 

\subsubsection{Linguistic context labeling}
Considering the same episodes for manual alignment of phrases previously described, we simultaneously labeled them on two linguistic context categories: topics and discourse genres. 
iLSU-T data was organized by topics according to the principal thematic axis, with the same conception as other studies in this field~\cite{albanie2021bbc, shen2024auslan}. iLSU-T covers a wide range of topics: weather, traffic, health, human rights, politics, social, culture, news, %safety and
security, laws and regulations, sports, and shows.
The term ``discourse genres'' refers to stereotyped forms of discourse, i.e., forms fixed by usage and repeated with relative stability in the same communicative situations. These discourse genres are frequently linked with a community of speakers in a particular context, for example, within a professional sphere. The genres share the same way of organizing information and the same set of linguistic resources, including register and phraseology~\cite{bajtin1952problema, ciapuscio1994tipos}. iLSU-T includes the following discourse genres: greetings and politeness formulae, reports, interviews, anecdotes and narratives, legal and normative procedures, debate and discussion, and argumentation.

\subsection{Dataset statistics}
\label{subsec:dataset_statistics}
\begin{figure}[t!]
    \centering
    \includegraphics[trim=2cm 4.8cm 6cm 0cm, clip=true, width=0.83\linewidth]{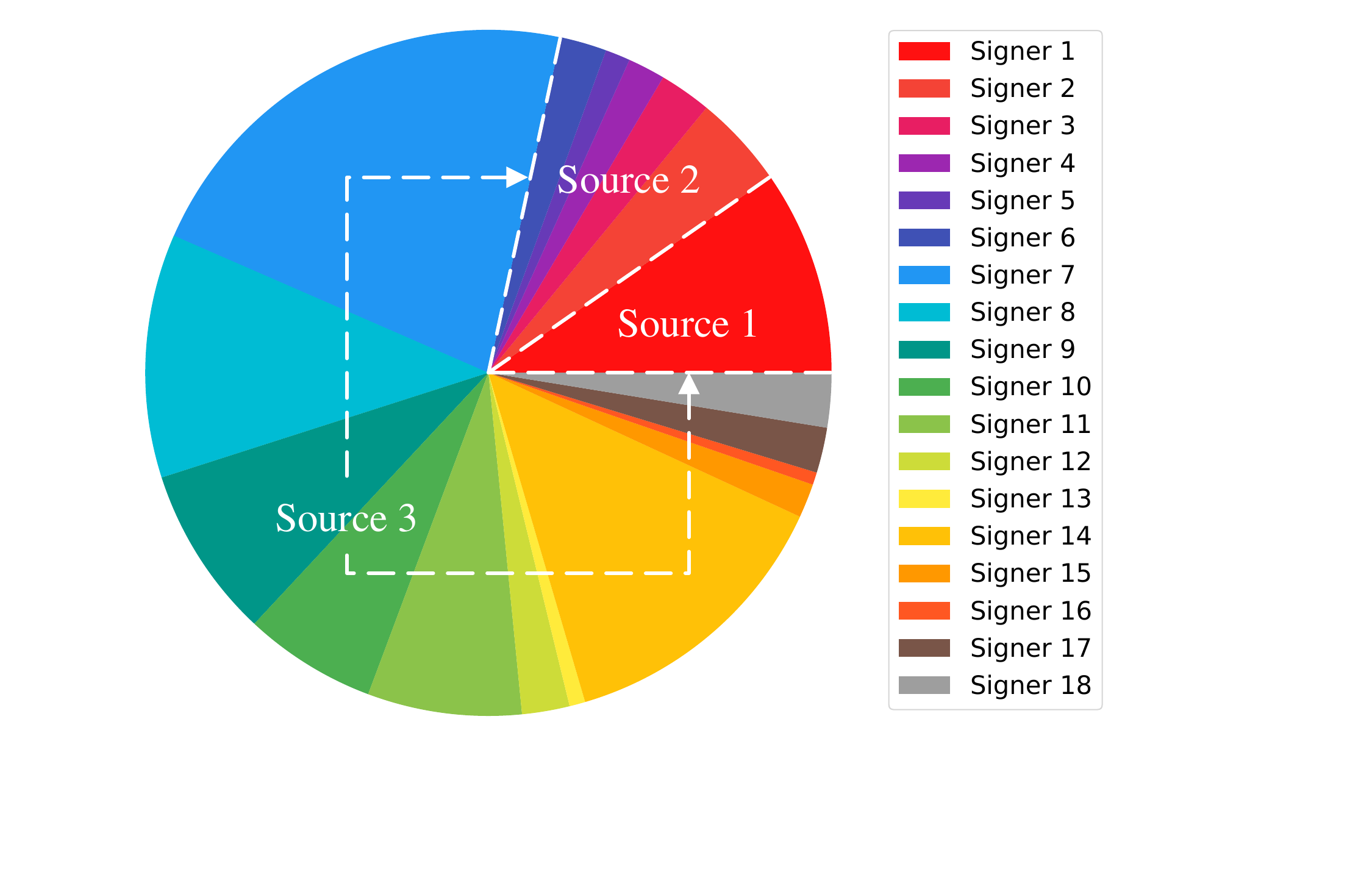}    
    \caption{Time duration distribution per signer and source in iLSU-T episodes.}
    \label{fig:signers_sources}
\end{figure}

\noindent
iLSU-T comprises 187.4 hours of RGB video interpreted in LSU and structured in 571 episodes with an average length of 19.7 minutes. Table~\ref{table:ilsut_statistics} shows its distribution between the 3 video sources. There are 18 signers in the whole dataset. The signers involved in each source are mutually exclusive. Fig.~\ref{fig:signers_sources} shows the distribution of the time duration per signer and source. Sources 1 and 2 have a frame rate of 25 fps for all episodes. Source 3 has two frame rate values, 25 and 30 fps, with a time duration proportion of about 3:7 for the lowest frame rate over the highest one.

\begin{table}[t!]
\caption{iLSU-T episodes statistics: RoI dimensions, time duration (in hours), vocabulary size, and number of signers per source. }
\label{table:ilsut_statistics}
\centering
\resizebox{0.49\textwidth}{!}
{\begin{tabular}{lllccc}
\toprule
Set/Subset   & RoI width & RoI height  & Duration\,[h] & Vocabulary & \#signers \\
\midrule
Whole dataset & 343.2$\pm$ 46.6  & 363.7$\pm$ 60.5 & 187.4 & 37k9  & 18   \\
\midrule
Source 1   & 331.9$\pm$ 2.7  & 312.9$\pm$ 2.6 & 18.1 & 12k3 & 1   \\
Source 2   & 246.6$\pm$ 11.8  & 240.1$\pm$ 3.9 & 22.4 & 14k1 & 5   \\
Source 3   & 362.2$\pm$ 27.8  & 393.2$\pm$ 29.0 & 146.9 & 29k9 & 12 \\
\bottomrule
\end{tabular}}
\end{table}

% \begin{figure}[b!]
%     \centering
%     \includegraphics[trim=2cm 4.8cm 6cm 0cm, clip=true, width=0.98\linewidth]{hours_per_signer.pdf}    
%     \caption{Time duration per signer in iLSU-T episodes.}
%     \label{fig:signers_sources}
% \end{figure}

% \begin{figure}[b!]
%     \centering
%     \includegraphics[trim=13cm 1cm 0cm 5cm, clip=true, width=0.98\linewidth]{hours_per_signer_source.pdf}    
%     \caption{Time duration distribution per signer and source in iLSU-T episodes.}
%     \label{fig:signers_sources}
% \end{figure}

% \begin{table}[b!]
% \caption{Time duration in hours per signer and source in iLSU-T episodes.}
% \label{table:signers_sources}
% \centering
% \resizebox{0.49\textwidth}{!}
% {\begin{tabular}{lccccccccccccccccccc}
% \toprule
% \multirow{2}{*}{Source} & \multicolumn{18}{c}{Signer ID}\\
%        & 1 & 2 & 3 & 4 & 5 & 6 & 7 & 8 & 9 & 10 & 11 & 12 & 13 & 14 & 15 & 16 & 17 & 18\\
% \midrule
% Source 1 & 18.1 &  &  &  &  &  &  &  &  & &  &  &  &  & &  &  & \\
% Source 2 &  & 8.2 & 4.6 & 3.3 & 2.2 & 4.1 &  &  &  & &  &  &  &  & &  &  & \\
% Source 3 &  &  &  &  &  &  & 40.9 & 21.5 & 15.2 & 11.7 & 13.6 & 4.2 & 1.4 & 25.4 & 3.0 & 1.1 & 4.0 & 4.8 \\
% \bottomrule
% \end{tabular}}
% \end{table}

% Comentar cuáles son las fuentes de datos de información y sus principales características, tanto en tamaño como en características del fondo y cantidad de señantes. Para esto se podria armar una tablita. Comentar que en general se tiene imágenes de media resolución, y que quizás no resultan útiles para la consideración de \textit{features} manuales, faciales o de \textit{mouthing}.

% Por cada fuente de información la tabla podría tener: 
% \begin{itemize}
%     \item Horas crudas.
%     \item Cantidad de señantes. Aclarar que los señantes son diferentes en todas fuentes de datos.
%     \item Bounding boxes: diagonal media y desviación estándar.
% \end{itemize}

\subsection{Dataset structure} \label{subsec::dataset_structure}
\noindent
As previously mentioned, the dataset comprises 571 episodes. % from three different sources. 
A unique text ID identifies episodes with the following information: media source, source file, time range in the source file, and signer~ID. The audio was automatically transcribed for each episode, with an independent timeline. Each episode includes the text track, as explained in Section~\ref{subsec:signlanguageprocessing}.
Sign language experts manually produce ground truth alignment of text and video tracks for over 20 hours of the iLSU-T dataset. These annotations are available for some episodes of each data source.

\subsection{Dataset license of use}
\noindent
% iLSU-T dataset was collected and published in the framework of an agreement between the National Telecommunications Directorate and the Universidad de la República del~Uruguay. The owners of the videos (the Uruguayan Parliament and both TV chains) explicitly agreed to use this data for research purposes.
iLSU-T dataset was collected and published in a collaboration between academia and media sources. The data is shared under a restricted use license (see data repository\footnote{\url{https://github.com/ariel-e-stassi/iLSU-T}.} for details) that allows its access and use for research and educational purposes.
%iLSU-T dataset is publicly available~on the~website~\textcolor{red}{URL (occulted for anonymization)}, only for research and educational purposes, after agreement of its Restricted Use License.

\section{EXPERIMENTS} \label{sec:experiments}
\noindent
This section describes experiments on the iLSU-T dataset using three state-of-the-art methods. We present each method and the experimental setup implemented.
 
\subsection{Methods}
\label{subsec:sota_methods}

\subsubsection{Sign Language Transformers (SLT)} In 2020, Camgoz et al.~\cite{camgoz2020sign} proposed a method based on the transformer architecture to simultaneously translate a sequence of video frames to sign language glosses and written language. For this purpose, the authors considered a joint loss function for simultaneous recognition and translation. % in a weighted fashion. 
The expression of the loss function is $\mathcal{L}=\lambda_R\, \mathcal{L}_R + \lambda_T\, \mathcal{L}_T$, where $\mathcal{L}_R$ and $\mathcal{L}_T$ are the recognition and translation loss, respectively. %~\cite{camgoz2020sign}. 
Because gloss annotations are unavailable in iLSU-T data, we considered $\lambda_R=0$ and $\lambda_T=1$.

\subsubsection{Stochastic Transformer Networks with Linear Competing Units: application to end-to-end SL translation (STLCU)} In 2021, Voskou et al.~\cite{voskou2021stochastic} proposed a method based on the transformer architecture with a novel scheme in the layer structure. Stochastically competing units replace the conventional ReLU activation functions, and the layer weights are fitted with a variational inference approach. This method includes a numerical compression strategy for the model weights. In this work, we use the STLCU model with full numerical precision.

\subsubsection{Gloss Attention for Gloss-Free Sign Language Translation (GASLT)} In 2023, Yin et al.~\cite{yin2023gloss} proposed a method for sign language translation from videos that takes into account textual information to solve the task by using the proximity of sentence embeddings. This notion of similarity computed for each pair of sentences of the dataset makes it possible to mitigate the lack of gloss supervision. In this work, we substituted the original BPE encoding with word encoding for the text, which performs better.

\subsection{SOTA methods on iLSU-T}
\subsubsection{Automatic video clipping}
Here, we refer to a video clip as a fragment that ideally corresponds to a text phrase. Automatic video clipping was performed using a methodology based on random delays. Similarly to Dal~Bianco et al.~\cite{dal2022lsa}, we consider a pre-delay and a post-delay between the beginning and ending times of the sign language video content and the beginning and ending times of each text phrase or utterance, respectively. 

\begin{figure}[t!]
    \centering
    \includegraphics[trim=1cm 0cm 4.8cm 2.5cm, clip=true, width=0.85\linewidth]{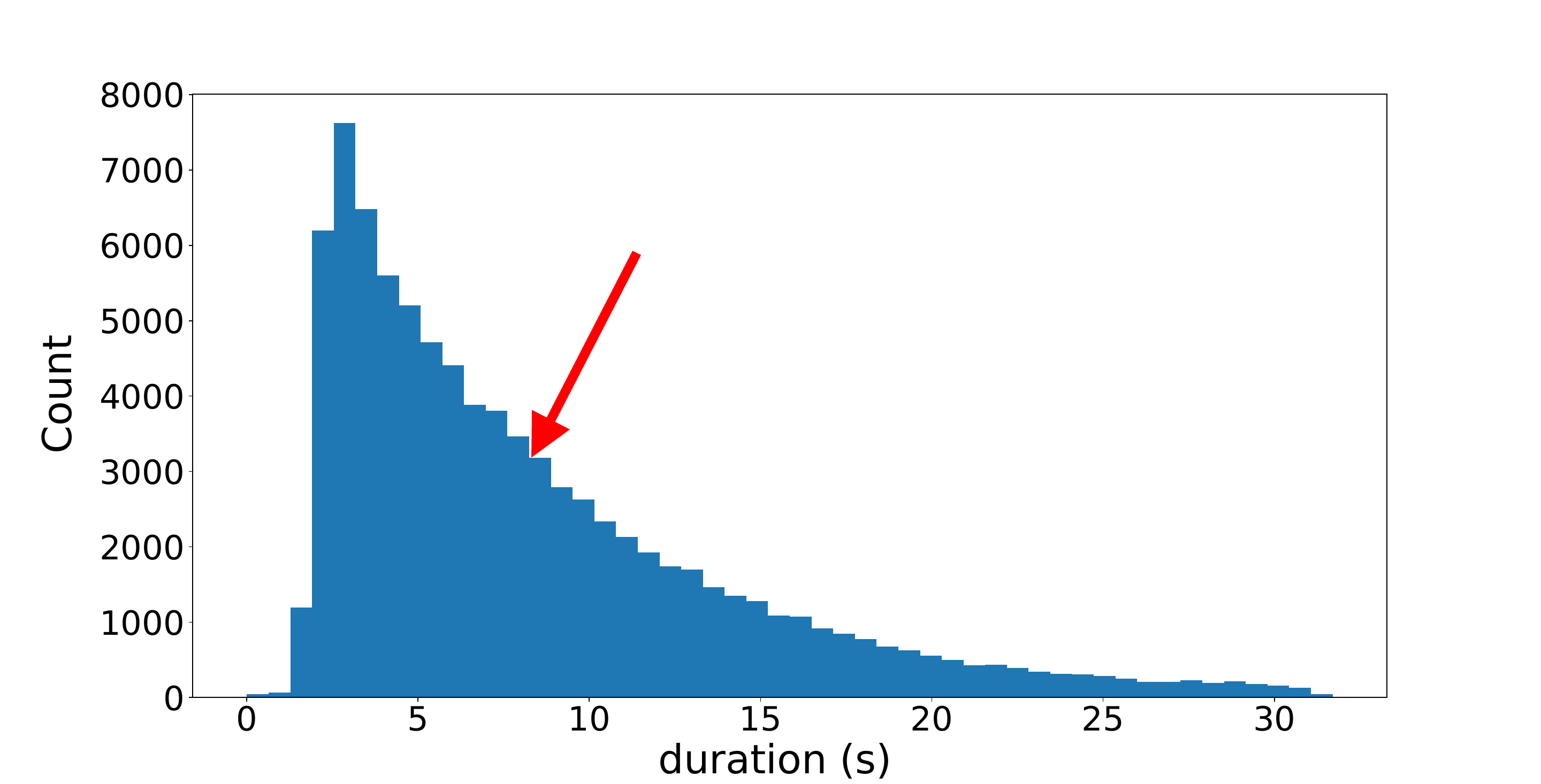} \caption{Histogram of video-clip durations for the whole iLSU-T dataset. The red arrow points to the average duration of 8.38 seconds.}
    \label{fig:video_clips_durations_hist}
\end{figure}

As a first approach to the ground-truth association between sign language video clips and their corresponding phrases or utterances, here we considered that pre-delay time~$t_1$ as well as post-delay time $t_2$ follows uniform distributions~\mbox{$t_1\sim\mathcal{U}(a_1,b_1)$} and $t_2\sim\mathcal{U} (a_2,b_2)$. %From this stochastic model and using visual inspection in random samples, 
We propose a first selection of \mbox{$[a_1, b_1, a_2, b_2]=[0.4,1.2,2.1,2.9]$} values for the whole dataset. These values were chosen by visual inspection from a random sample of episodes trying to compose video clips containing the complete sign language phrase %or sentence 
associated with the text sentence. With this approach, 86550 video clips were obtained with an average duration of 8.38 seconds and a standard deviation of 5.95. Fig.~\ref{fig:video_clips_durations_hist} shows a histogram of video-clip durations for the whole dataset. Note that some clips are more than 20 seconds long. The video clips can overlap, so the total duration of all the video clips is 201.52 hours. Despite the temporal overlapping between the visual content of two consecutive fragments of an episode, the text content of each video clip is \textit{a priori} independent from the others.

\subsubsection{Datasets and data splitting}
\noindent
Four iLSU-T data configurations are proposed for training and testing. The first is to consider a random splitting of all video clips, hereafter referred to as the whole dataset. Based on the data sources, additional data subsets are proposed to study the performance of the methods in slightly more controlled scenarios. Hereafter, these subsets will be referred to as Source~1, Source~2, and Source~3, respectively. For each data configuration, video clips were randomly split into train, validation, and test sets, considering a proportion of 0.8, 0.1, and 0.1, respectively. 

\subsection{Reproduction details}
\label{subsec:repro_details}
\subsubsection{I3D visual features}
\noindent
The tested methods were fed with video visual features 
provided by feature extractors previously trained. Two feature extractors were considered based on the I3D architecture originally proposed in~\cite{carreira2016human}. Each video clip input is reorganized as a sequence of overlapped sub-video clips. This sequence is determined by the window width and stride defined in~\cite{li2020tspnet}. 
We used the official implementation of the method TSPNet~\cite{li2020tspnet}, with a window width value of 8 frames and a stride of 2 frames. The I3D network was used as a frozen feature extractor by considering existing pre-trained weights for two sign language tasks. The first one, I3D-ASL2k, is a model trained for 
sign language recognition or classification over the 2000 isolated signs of the American Sign Language WLASL dataset~\cite{li2020transferring}. The weights of this model were obtained from \url{https://github.com/verashira/TSPNet}. The second one, I3D-BSL5k, is a model trained for 
temporal sign localization considering attention localizations, mouthing, and dictionary annotations (M+D+A model) over a vocabulary of 5383 words of the British Sign Language~\cite{varol_read_2021}. The weights of this model were obtained from \url{https://www.robots.ox.ac.uk/~vgg/research/bslattend/}.

\subsubsection{Sentence-embedding similarities in GASLT method}
\noindent
For the reproduction of the GASLT method, a semantic similarity matrix for each dataset was calculated. This matrix comprises the cosine similarity between each pair of sentences in the datasets. We followed the procedure presented in the official repository of the method~\cite{yin2023gloss}. However, given its sizes, it was necessary to compute this matrix by parts in two of the four data %whole dataset and source three 
configurations. To this aim, the GASLT model was slightly modified to reconstruct the similarity matrix from its parts internally.

\subsubsection{Training details}
\noindent
All the SOTA methods were trained for a maximum of 100 epochs with a batch size of 128~samples, except for the GASLT method on the whole dataset, which used a 64-sample batch size due to RAM restrictions. We explore some variations from the default training configuration for each considered method. %For example, patience was set to 15 epochs. 
Configuration files for each method are included in the iLSU-T repository.
% For the GASLT method, the similarity loss weight hyperparameter value was set to 5.

\subsection{Evaluation metrics}
\noindent
We used two classical metrics for the quantitative evaluation of automatic translations: $\mathrm{ROUGE-L}$ and $\mathrm{BLEU-}N$. For the $\mathrm{ROUGE-L}$ metric, we must consider the longest common sequence between two sequences of words or tokens. Let be $X$, a reference translation of length $r$, and $Y$, a candidate translation of length $c$. We denote as $\mathrm{LCS}(X,Y)$ the longest common sequence between $X$~and~$Y$. Then, 
\begin{equation}
\label{eq:rouge}
\mathrm{ROUGE-L} = \frac{(1+\beta^2)R_{\mathrm{LCS}}P_{\mathrm{LCS}}}{R_{\mathrm{LCS}}+\beta^2 P_{\mathrm{LCS}}},
\end{equation}
\noindent
with $R_{\mathrm{LCS}}=\frac{\mathrm{LCS}(X,Y)}{r}$ and $P_{\mathrm{LCS}}=\frac{\mathrm{LCS}(X,Y)}{c}$~\cite{lin2004automatic}. In this work, we use $\beta=1.2$ as in the official implementation of the selected methods~\cite{camgoz2020sign, voskou2021stochastic, yin2023gloss}. 

For $\mathrm{BLEU-}N$ metric we must consider the concept of \textit{modified precision score}, denoted as $p_n$ and defined as~\cite{papineni2002bleu}:
\begin{equation}
   p_n = \frac{\sum_{Y} \sum_{n\mathrm{-gram}\in\, Y} \mathrm{\#}_{clip} (n\mathrm{-gram})}{\sum_{Y} \sum_{n\mathrm{-gram}\in\, Y} \mathrm{\#}(n\mathrm{-gram})},
\end{equation}
\noindent
where $n\mathrm{-gram}$ is a ``sequence of $n$ words.''~\cite{jm3}, and
\begin{equation}
    \label{eq:count_clip}
    \mathrm{\#}_{clip}(n\mathrm{-gram}) = \mathrm{min}\{\mathrm{\#}(n\mathrm{-gram}), \max_\mathcal{X}\{\mathrm{\#}(n\mathrm{-gram})\}\}.    
\end{equation}
\noindent
In Equation~\ref{eq:count_clip}, $\#$ represents the counting operation and $\mathcal{X}$ a set of reference sentences for a given candidate sentence~$Y$. Then, $\mathrm{BLEU-}N$ is computed as~\cite{papineni2002bleu}: 
\begin{equation}
\label{eq:bleu}
\mathrm{BLEU-}N = BP \cdot\, \exp{\left(\sum_{n=1}^{N} w_n \log{p_n}\right)},
\end{equation}
\noindent
where $w_n=1/N$ and $BP=\min\left\{1,\exp{(1-r/c)}\right\}$ is a penalty for brevity of the translation. In this work we use SacreBLEU, a standardized tool for computing reproducible and comparable BLEU scores~\cite{post_2018_call}.

\begin{table*}[h!]
\caption{Baseline on the whole iLSU-T dataset with Selected SOTA Methods (\colorbox{red!15}{best value}, \colorbox{blue!15}{second-best value}).}
\centering
\resizebox{\textwidth}{!}
{
\begin{tabular}{lccccccccccc}
\toprule
\multirow{2}{*}{Method}  & {Visual feature} & \multicolumn{5}{c}{DEV}                                            & \multicolumn{5}{c}{TEST} \\
                &  extraction    & BLEU-1$\,\uparrow$      & BLEU-2$\,\uparrow$      & BLEU-3$\,\uparrow$      & BLEU-4$\,\uparrow$      &  ROUGE-L$\,\uparrow$      & BLEU-1$\,\uparrow$      & BLEU-2$\,\uparrow$     & BLEU-3$\,\uparrow$      & BLEU-4$\,\uparrow$      &  ROUGE-L$\,\uparrow$  \\
\midrule
\multirow{2}{*}{SLT~\cite{camgoz2020sign}}      &  I3D-ASL2k    &  15.10 & 5.55 & 2.30 & 1.24 & 11.25 & 15.09 & 5.46 & 2.14 & 1.10 & 11.40\\
      &   I3D-BSL5k    & 17.98 & 6.04 & 2.42 & {1.31} & {11.42} & 18.00 & 6.06 & 2.34 & {1.24} & {11.57}\\
\midrule
\multirow{2}{*}{STLCU~\cite{voskou2021stochastic}}  &  I3D-ASL2k  & 17.46 & 8.10 & 4.90 & \colorbox{red!15}{3.45} & \colorbox{red!15}{14.65} & 17.69 & 8.17 & 4.92 & \colorbox{red!15}{3.43} & \colorbox{red!15}{14.86} \\
                  &   I3D-BSL5k      & 14.81   & 5.01    &    2.15         &    1.16         &  11.27  & 14.76 & 4.98    &  2.04  &  1.03 & 11.45     \\
\midrule
\multirow{2}{*}{GASLT~\cite{yin2023gloss}}    & I3D-ASL2k  &  15.32  &  5.60   &   2.54    &   \colorbox{blue!15}{1.37}   &    \colorbox{blue!15}{11.53}   &    15.61         &    5.64         &    2.49         &   \colorbox{blue!15}{1.29}          &  \colorbox{blue!15}{11.57}   \\
   &   I3D-BSL5k & 13.29    &     4.78        &    2.18         &   1.14          &   10.15         &   13.26           &    4.71         &    2.06         &     1.03        &     10.09    \\
\bottomrule
\end{tabular}
}
\label{table:baseline_wholeds}
\end{table*}

\begin{table*}[h!]
\caption{Baseline on Source-1 iLSU-T videos with Selected SOTA Methods (\colorbox{red!15}{best value}, \colorbox{blue!15}{second-best value})}
\centering
\resizebox{\textwidth}{!}
{
\begin{tabular}{lccccccccccc}
\toprule
\multirow{2}{*}{Method}  & {Visual feature} & \multicolumn{5}{c}{DEV}                                            & \multicolumn{5}{c}{TEST} \\
                &  extraction    & BLEU-1$\,\uparrow$      & BLEU-2$\,\uparrow$      & BLEU-3$\,\uparrow$      & BLEU-4$\,\uparrow$      &  ROUGE-L$\,\uparrow$      & BLEU-1$\,\uparrow$      & BLEU-2$\,\uparrow$     & BLEU-3$\,\uparrow$      & BLEU-4$\,\uparrow$      &  ROUGE-L$\,\uparrow$  \\
\midrule
\multirow{2}{*}{SLT~\cite{camgoz2020sign}}      &  I3D-ASL2k      & 14.58 & 4.74 & 1.51 & 0.72 & \colorbox{red!15}{11.19} & 14.50 & 4.71 & 1.60 & {0.71} & \colorbox{red!15}{10.97}\\
                                                &   I3D-BSL5k    &  13.35 & 3.77 & 1.19 & 0.57 & \colorbox{blue!15}{9.44} & 13.22 & 3.66 & 1.16 & 0.53 & \colorbox{blue!15}{9.26}\\
\midrule
\multirow{2}{*}{STLCU~\cite{voskou2021stochastic}}  &  I3D-ASL2k        & 12.99 & 4.41 & 2.12 & \colorbox{red!15}{1.23} & 9.04 & 11.97 & 3.77 & 1.59 & \colorbox{blue!15}{0.88} & 8.72\\
                  &   I3D-BSL5k      &  11.53 & 3.45 & 1.46 & \colorbox{blue!15}{0.79} & 7.70 & 11.27 & 3.19 & 1.58 & \colorbox{red!15}{1.01} & 7.44\\
\midrule
\multirow{2}{*}{GASLT~\cite{yin2023gloss}}  & I3D-ASL2k        &  11.16   & 4.00            &    1.51         &     0.67        & 9.33            &    10.81         &   3.67          &     1.23        &     0.31        &   8.96      \\
   &   I3D-BSL5k      & 11.56 &  4.26           &   1.77          &   0.77          &  9.01           &   11.69          &   4.03          &   1.44          &    0.44         &  9.13   \\
\bottomrule
\end{tabular}
}
\label{table:source1}
\end{table*}

\begin{table*}[h!]
\caption{Baseline on Source-2 iLSU-T videos with Selected SOTA Methods (\colorbox{red!15}{best value}, \colorbox{blue!15}{second-best value})}
\centering
\resizebox{\textwidth}{!}
{
\begin{tabular}{lccccccccccc}
\toprule
\multirow{2}{*}{Method}  & {Visual feature} & \multicolumn{5}{c}{DEV}                                            & \multicolumn{5}{c}{TEST} \\
                &  extraction    & BLEU-1$\,\uparrow$      & BLEU-2$\,\uparrow$      & BLEU-3$\,\uparrow$      & BLEU-4$\,\uparrow$      &  ROUGE-L$\,\uparrow$      & BLEU-1$\,\uparrow$      & BLEU-2$\,\uparrow$     & BLEU-3$\,\uparrow$      & BLEU-4$\,\uparrow$      &  ROUGE-L$\,\uparrow$  \\
\midrule
\multirow{2}{*}{SLT~\cite{camgoz2020sign}}      &  I3D-ASL2k      & 15.55 & 5.13 & 1.71 & 0.79 & \colorbox{blue!15}{10.84} & 15.32 & 5.00 & 1.39 & 0.55 & {10.58}\\
                                                &   I3D-BSL5k    &   10.74 & 3.81 & 1.14 & 0.43 & {10.59} & 10.88 & 3.72 & 0.99 & 0.41 & 10.21\\
\midrule
\multirow{2}{*}{STLCU~\cite{voskou2021stochastic}}  &  I3D-ASL2k    &   16.57 & 5.78 & 2.72 & \colorbox{red!15}{1.69} & 10.62 & 16.12 & 5.34 & 2.25 & \colorbox{red!15}{1.30} & \colorbox{blue!15}{10.72}\\
                  &   I3D-BSL5k      & 13.81 & 4.46 & 2.14 & \colorbox{blue!15}{1.37} & 8.80 & 13.81 & 4.04 & 1.70 & \colorbox{blue!15}{1.01} & 8.70\\
\midrule
\multirow{2}{*}{GASLT~\cite{yin2023gloss}}    & I3D-ASL2k   &  16.49  &   5.76    &  2.20   &   0.90   &   \colorbox{red!15}{11.20}   &   15.65     & 5.35   &   1.97   &  0.82  & \colorbox{red!15}{10.74}     \\
                                              & I3D-BSL5k   &  16.00  &    5.45         &    2.06         &   0.96          &    10.47         &   15.35          &   5.23          &    1.79         &      0.75       &    9.72 \\
\bottomrule
\end{tabular}
}
\label{table:source2}
\end{table*}

\begin{table*}[h!]
\caption{Baseline on Source-3 iLSU-T videos with Selected SOTA Methods (\colorbox{red!15}{best value}, \colorbox{blue!15}{second-best value})}
\centering
\resizebox{\textwidth}{!}
{
\begin{tabular}{lccccccccccc}
\toprule
\multirow{2}{*}{Method}  & {Visual feature} & \multicolumn{5}{c}{DEV}                                            & \multicolumn{5}{c}{TEST} \\
                &  extraction    & BLEU-1$\,\uparrow$      & BLEU-2$\,\uparrow$      & BLEU-3$\,\uparrow$      & BLEU-4$\,\uparrow$      &  ROUGE-L$\,\uparrow$      & BLEU-1$\,\uparrow$      & BLEU-2$\,\uparrow$     & BLEU-3$\,\uparrow$      & BLEU-4$\,\uparrow$      &  ROUGE-L$\,\uparrow$  \\
\midrule
\multirow{2}{*}{SLT~\cite{camgoz2020sign}}      &  I3D-ASL2k      & 14.63 & 5.98 & 3.53 & 2.48 & 12.93 & 14.68 & 5.79 & 3.31 & 2.26 & 13.03\\
      &   I3D-BSL5k    &   15.90 & 5.50 & 2.41 & 1.35 & 11.96 & 15.83 & 5.26 & 2.10 & 1.07 & 11.78\\
\midrule
\multirow{2}{*}{STLCU~\cite{voskou2021stochastic}}  &  I3D-ASL2k        &   18.74 & 9.00 & 5.65 & \colorbox{red!15}{4.08} & \colorbox{red!15}{16.18} & 18.44 & 8.75 & 5.40 & \colorbox{red!15}{3.82} & \colorbox{red!15}{16.05}\\
                  &   I3D-BSL5k      &  13.54 & 4.95 & 2.53 & {1.54} & {11.80} & 13.45 & 4.55 & 2.15 & {1.18} & {11.66}\\
\midrule
\multirow{2}{*}{GASLT~\cite{yin2023gloss}}    & I3D-ASL2k  &   19.29  &  8.08           &   4.34          &   \colorbox{blue!15}{2.73}          &   \colorbox{blue!15}{15.12}          &    19.04         &   7.65          &    3.89         &       \colorbox{blue!15}{2.34}      &   \colorbox{blue!15}{14.80}      \\ 
   &   I3D-BSL5k      &  16.72   &  6.33           &   3.03          &  1.69           &    12.43         &   16.39          &    5.98         & 2.76            &   1.47          &   12.20      \\
\bottomrule
\end{tabular}
}
\label{table:source3}
\end{table*}
% In this work it is used SacreBLEU implementation for computing BLEU-$N$ metric. This implementation allows a more reliable comparison of the metrics between different 

\section{RESULTS} \label{sec:results}
\noindent
This section presents the results obtained by applying the selected methods to the four data configurations previously described. Tables~\ref{table:baseline_wholeds},~\ref{table:source1},~\ref{table:source2}, and~\ref{table:source3} show the \mbox{BLEU-$N$} metrics and ROUGE-L for the validation (DEV) and test (TEST) sets obtained in each of the splits. %For metrics expressions please see Equations~\ref{eq:rouge} and~\ref{eq:bleu}. 
Table~\ref{table:translation_examples} shows translation examples using the three considered methods.

\begin{table*}[h!]
\caption{Qualitative test examples with I3D-ASL2k visual features' extractor. Models trained and tested on Source-3 data.}
\centering
\resizebox{0.98\textwidth}{!}
{
\begin{tabular}{r p{0.40\linewidth} rrrrrr}
\toprule
Method & Selected example & BLEU-1$\,\uparrow$ & BLEU-2$\,\uparrow$ & BLEU-3$\,\uparrow$ & BLEU-4$\,\uparrow$ & ROUGE-L$\,\uparrow$ & BERTScore$\,\uparrow$ \\
\midrule
Reference & \textit{continuando con la lista de oradores, tiene la palabra el senador adrián peña.} & & & & & \\
  SLT & tiene la palabra el senador doménech. & 25.95 & 25.43 & 24.72 & 23.66 & 49.35 & 0.793 \\
SCULT & tiene la palabra la senadora de la aventura. & 33.45 & 22.61 & 16.59 & 0.00 & 27.39 & 0.772 \\
GASLT & continuando con la lista de oradores, tiene la palabra el senador mieres. & 84.34 & 83.99 & 83.59 & 83.14 & 87.36 & 0.947 \\
\midrule
Reference & \textit{bueno, muchas gracias, señora presidenta.} & & & & & \\
  SLT & gracias, señora presidenta. & 51.34 & 51.34 & 51.34 & 0.00 & 71.76 & 0.891 \\
SCULT & muchas gracias, señora presidenta. & 77.88 & 77.88 & 77.88 & 77.88 & 87.14 &0.929\\
GASLT & gracias, señora presidenta. & 51.34 & 51.34 & 51.34 & 0.00 & 71.76 & 0.891 \\
\midrule
Reference & \textit{24 en 24.} & & & & & \\
  SLT & 23 en 23. & 33.33 & 0.00 & 0.00 & 0.00 & 33.33 & 0.960 \\
SCULT & 23 en 25. & 33.33 & 0.00 & 0.00 & 0.00 & 33.33 & 0.892\\
GASLT & 23 en 26. & 33.33 & 0.00 & 0.00 & 0.00 & 33.33 & 0.882 \\
\midrule
Reference & \textit{continuando con el debate, tiene la palabra el senador mieres.} & & & & & \\
  SLT & a la sesión de la comisión de asuntos laborales y seguridad social. & 8.33 & 0.00 & 0.00 & 0.00 & 9.24 & 0.699\\
SCULT & tiene la palabra el senador germán coutinho. & 46.53 & 44.95 & 42.91 & 40.05 & 57.01 &  0.773\\
GASLT & continuando con la lista de oradores, tiene la palabra el senador bordaberri. & 33.33 & 0.00 & 0.00 & 0.00 & 64.70 & 0.879 \\
\midrule
Reference & \textit{vamos a votar la solicitud de licencia leída, se está votando.} & & & & & \\
  SLT & tiene la palabra el senador martínez huelmo. & 8.07 & 0.00 & 0.00 & 0.00 & 10.68 & 0.650\\
SCULT & se va a votar la solicitud de licencia, se está votando. & 72.73 & 66.06 & 57.88 & 46.92 & 72.73 & 0.915\\
GASLT & gracias, señor senador. & 0.00 & 0.00 & 0.00 & 0.00 & 0.00 & 0.682\\
\midrule
Reference & \textit{a ese cuenta de otra solicitud y licencia llegada a la mesa.} & & & & & \\
  SLT & vamos a votar la licencia solicitada. & 18.39 & 0.00 & 0.00 & 0.00 & 20.96 & 0.729\\
SCULT & gracias, señora senadora. & 0.00 & 0.00 & 0.00 & 0.00 & 0.00 & 0.682\\
GASLT & vamos a votar la solicitud de licencia llegada a la mesa. & 66.41 & 49.25 & 41.95 & 36.03 & 60.40 &0.823\\
\bottomrule
\end{tabular}
}
\label{table:translation_examples}
\end{table*}

% DISCUSION
% \textcolor{red}{Reasons for performance on the different data splits. Vocabulary size, amount of data.
% Comentar mejores casos, comentar peores casos. 
% Tomar como referencia los resultados de GASLT sobre SP10 y CSLDaily. Resultados de Auslan-Daily.
% Coreference Resolution (planteado en discusión de Auslan-Daily), entiendo que se refiere al hecho de emplear cierta región del espacio de señado para cierto concepto... Vocabulary size
% Video resolution
% Comentar el problema de puntuar en la oralidad (Punctuation prediction in ASR systems), a su vez, comentar el tema de posibles errores de transcripción del texto (como un problema adicional).}

% ACA

\section{DISCUSSION}
\label{sec:discussion}
\noindent
Tables~\ref{table:baseline_wholeds}~to~\ref{table:source3} show that there are significant differences in the performance depending on the considered combination of datasets and methods. Source-3 iLSU-T presented the best results in general, regardless of the process. This behavior could be explained by the fact that this source has more data and includes duplicate text phrases between its different internal splits, i.e., train, validation, and test sets. The best performance obtained on this dataset was 3.82 for \mbox{BLEU-4} and 16.05 for ROUGE-L. The better performance of the methods on Source-3 configuration is not due to a higher spatial resolution nor a higher frame rate compared to the other two sources' datasets. Before the extraction of visual features by the I3D network, frames are rescaled to a size of 224$\times$224 pixels regardless of the original resolution. 

Concerning the frame rate differences, an experiment was carried out to evaluate the effect of the frame rate on the video content representation. As presented in~Section~\ref{subsec:dataset_statistics}, Source-3 videos have native frame rates of 25 and 30 fps. We resampled the 30-fps video clips to 25-fps clips before visual feature extraction. Then, the results were compared with and without resampling using the same I3D-ASL2k with fixed window width and stride values --8 and 2, respectively-- and the same data splitting for training, validation, and testing. We ran 50 epochs for training. Table~\ref{table:25fps_versus_mixed} shows that the tested approaches are robust to small changes in the video frame rate, handling 25 and 30 fps without further training.

\begin{table}[b!]
\caption{Effect of Source-3 videos frame rate on SCULT performance.}
\centering
\resizebox{0.48\textwidth}{!}{
\begin{tabular}{cccccc}
\toprule
frame rate & BLEU-1$\,\uparrow$ & BLEU-2$\,\uparrow$ & BLEU-3$\,\uparrow$ & BLEU-4$\,\uparrow$ & ROUGE-L$\,\uparrow$\\
\midrule
25 \& 30 fps & 18.81 & 9.01 & 5.51 & 3.88 & 16.41 \\
only 25 fps & 18.37 & 8.78 & 5.40 & 3.80 & 16.15 \\
\bottomrule
\end{tabular}
}
\label{table:25fps_versus_mixed}
\end{table}

Although these results are still far from ideal values, it is worth noting that translation results for other datasets of similar complexity in terms of the number of samples and vocabulary size are in the same order. Let's take two datasets, OpenASL and How2Sign, as examples. In Table~\ref{table:slt_datasets}, it can be seen that both datasets have a similar size to Source-3~\mbox{iLSU-T}. For OpenASL~\cite{shi2022open}, the best \mbox{BLEU-4} and ROUGE-L metrics obtained were respectively 6.57 and 21.02 with an approach proposed by the authors, based on a multi-stream translation system fed by global, handshape, and mouthing feature sequences. The authors highlight remarkable differences in the translation performance depending on the presence or absence of duplicate phrases, i.e., whether text phrases are present in the train and test sets. % SERIA BUENO MOSTRAR ALGUNA PRUEBA ACA de DESEMPEÑO sobre frases duplicadas, o no duplicadas.
For the How2Sign dataset~\cite{Duarte_CVPR2021}, a study that explores translation performance obtains a \mbox{BLEU-4} metric of 8.02~\cite{tarres2023sign}. 

\begin{figure}[b!]
    \centering
    \includegraphics[width=0.98\linewidth]{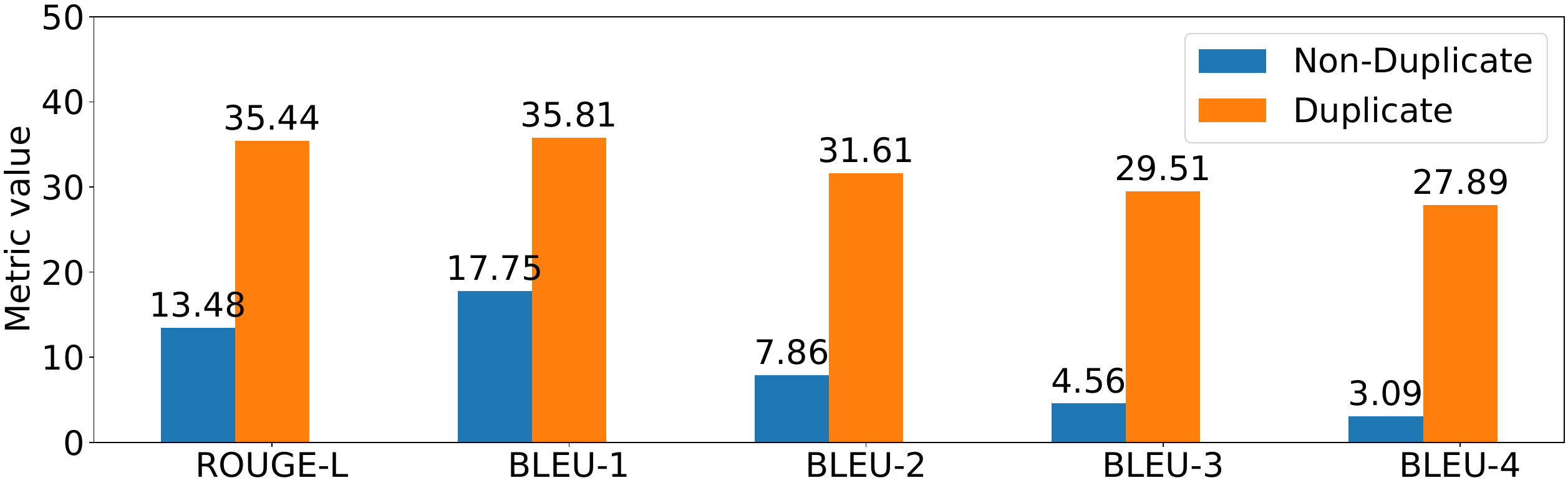} \caption{Effects of duplicate phrases on translation performance.}
    \label{fig:dup_nondup_phrases}
\end{figure}

A qualitative analysis of some selected examples shows that the studied models produce, in some cases, translations with sense. There are phrases like ``dese cuenta de otra solicitud de licencia llegada a la mesa.'' (``Register another license application.'') or ``se está votando.'' (``voting is underway'') for which automatic translations are almost perfect. This is expected due to the high frequency of these expressions in the Parliament sessions, where such phrases are part of the daily protocol of this legislative body. As in~\cite{shi2022open}, Fig.~\ref{fig:dup_nondup_phrases} shows the translation performance of SCULT I3D-ASL2k over two subsets of the Source-3 test set, one composed of 736 duplicate phrases and the other composed of 5552 non-duplicate phrases, both w.r.t. the phrases of the training set. 

Table~\ref{table:translation_examples} illustrates some example translations and their corresponding metrics. Note that the last column of the table corresponds to BERTScore~\cite{zhang2019bertscore} for Spanish. This metric ranges from 0 to 1 and captures each method's semantic similarity between a reference and candidate sentences. For example, let's see the phrase ``24 en 24.'' (``24 in 24''), concerning a vote in the Chamber. All methods show a poor performance in the sense of \mbox{BLEU-$N$} and ROUGE-L metrics, but not in the BERTScore metric, which measures a high degree of correspondence between the sentences. The three candidate phrases refer to a numerical proportion as the reference does. Moreover, BERTScore correctly captures that the SLT method proposes the same numbers in the proportion, even if they differ from the reference, giving a higher score for this method. This example shows the limitations of the \mbox{BLEU-$N$} and ROUGE-L metrics in capturing the difference in meaning between a reference and a candidate phrase. Those metrics count the number of equal words differently; in this case, only one word is correct. Even more, when the number of words in a phrase is lower than $N$, the BLEU-$N$ score is 0, as it is impossible to find $N$ correct words. BLEU-$N$ is used to characterize the translation performance on a corpus, but this issue hinders its global value for a dataset with several short phrases.

%Following the methodology, 
We can highlight some limitations on \mbox{iLSU-T} video clips. As seen in Table~\ref{table:translation_examples}, there are some challenges in automatically generating video clips at two levels: text and video. 

Two significant problems can be noticed in text generation. First, sentences like ``a ese cuenta de otra solicitud y licencia llegada a la mesa'' have not been correctly transcribed by WhisperX. Second, like any automatic speech recognition system, WhisperX has limitations on punctuation prediction. This is an open problem frequently associated with oral discourse~\cite{Zhou_Punctuation_2022}. The correct generation of video clips partially depends on the proper punctuation of the sentences. 

For video content, different problems arise: the interpreters omit performing some signs, switch to fingerspelling to refer to proper nouns, and use different signs for each one with their corresponding significance. Sign omission can be conceptualized as a data augmentation phenomenon in the best case, but clearly, it is a language aspect that is hard to control. Since the data is not labeled about fingerspelling, the trained models are not explicitly supervised concerning the switch between different signing modalities. Finally, another sign language particularity is called \textit{coreference resolution}, which has been discussed in the context of automatic sign language translation by Shen et al.~\cite{shen2024auslan}. Coreference resolution refers to using a specific region of the signing space to refer to an object previously introduced. The considered methods are trained and tested on isolated phrases; hence, the methods tested in the present study do not account for this phenomenon.

Finally, regarding the alignment at the sentence level, it is important to highlight that iLSU-T is composed of a series of episodes of approximately 20 minutes in length, which are split into shorter video clips to create batches to feed the neural network models, as explained in Section~\ref{subsec:signlanguageprocessing}. In this work, video clips' conformation is based on empirically adjusted random delays. This practical strategy allows for a first approximation of using iLSU-T data to train and test three selected translation methods. %As previously explained, iLSU-T comprises a series of episodes that must be procbibessed to obtain suitable datasets for automatic translation. 
The problem of automatic sign language segmentation and automatic alignment between the text and the video content is an open problem~\cite{Bull_2021_ICCV,bartoli_automatic_2020}. 

\section{CONCLUSIONS AND FUTURE WORKS}
\label{sec:conclusions}
\noindent
In this work, we introduced iLSU-T, a new dataset for automatic translation of interpreted Uruguayan Sign Language. A reproducible pipeline is also presented for processing raw data and obtaining the dataset episodes. The statistics reflect a dataset with diverse topics and numerous signers with a video quality similar to one of the most popular benchmarks in the field, i.e., Phoenix2014T. The state-of-the-art methods tested are exclusively based on visual features directly extracted from the video. BLEU-$N$ and ROUGE-L metrics values show that iLSU-T presents significant challenges when performing automatic translations to Spanish, the written or spoken language commonly used by hearing people in Uruguay. 

Two major fronts appear as future lines of work: methods and data. Regarding the methods, we must study the limitations of each considered method, focusing on the effects of the alignment between the text and the video track of the interpreted videos. Remarkably, the three considered methods are exclusively based on visual feature inputs. In this sense, we will explore strategies based on skeleton data, either using one-stream or multi-modal approaches. Concerning the data, it is crucial to enrich the annotations to conduct controlled experiments. For example, in this paper, we only considered visual features derived globally from each video RoI, i.e., the bounding box where the sign language interpreter appears. We do not consider features associated with the activity of the hands, face, or lips as is often done in the sign language translation field~\cite{shi2022open, varol_read_2021}. Secondly, it is important to consider the effect of various aspects of the text. Among others, it is necessary to conduct experiments that consider the length of the phrases and the amount of text phrase duplication between the train and test sets. Finally, enriching the text annotations by considering multiple references to evaluate the metrics is important. Generative text tools could be used for this purpose, which take an input sentence and provide multiple alternatives according to different similarity criteria based on semantics and language expressions. 

%%%%%%%%%%%%%%%%%%%%%%%%%%%%%%%%%%%%%%%%%%%%%%%%%%%%%%%%%%%%%%%%%%%%%%%%%%%%%%%%
\section{ACKNOWLEDGMENTS}
\noindent
iLSU-T was partially supported by a CAP--UdelaR scholarship, Uruguay. Some of the experiments were carried out using ClusterUY. We acknowledge DiNaTel Uruguay for providing us with the raw data, the NICA--UdelaR team for fruitful interdisciplinary discussions, and G. Gómez and F. Lecumberry for their website assistance.

%%%%%%%%%%%%%%%%%%%%%%%%%%%%%%%%%%%%%%%%%%%%%%%%%%%%%%%%%%%%%%%%%%%%%%%%%%%%%%%%
\section*{ETHICAL IMPACT STATEMENT}
\noindent
The main contribution of our database is the alignment and preprocessing of the video and signed sequences; since the video data is taken from open TV broadcasting, the privacy risks are assessed as low and very low (this assessment was provided by the ethics committee that reviewed and exempted the work). Potential risks of misuse include users exploiting the data for unethical purposes or developing malicious algorithms. To mitigate these risks, we define rules for using the iLSU-T, with open but controlled access to the data stated in the restricted use license. 

%The translation methods can have biases %discussed in this paper have poor performances on Uruguayan Sign Language and then are still unreliable for practical applications. Using these systems for automatic transcription from sign language videos 
%that could cause communication problems between signers and interlocutors who receive the message in text format. To mitigate this risk, we adopt two combined strategies: limiting the response capabilities of the system when certainty levels of automatic translation are below some ad hoc thresholds and involving real users for perceptual feedback on the quality of the developed solutions. Finally, it is crucial to highlight the importance of interdisciplinary work, specially with linguistic specialists. 

% This year, at FG 2025, we will introduce a new requirement that authors submit an Ethical Impact Statement as part of the submission process. (Note that this requirement applies to all short and long papers submitted to the main track. Each special track may have its own rules.) To support authors in meeting this requirement and reviewers in assessing it, we have prepared this document with general guidelines, a checklist for authors and reviewers to complete, and answers to some frequently asked questions. As this is a new policy, we welcome questions and feedback as we refine it and develop a shared understanding.
%%%%%%%%%%%%%%%%%%%%%%%%%%%%%%%%%%%%%%%%%%%%%%%%%%%%%%%%%%%%%%%%%%%%%%%%%%%%%%%%

{\small
\bibliographystyle{ieee}
\bibliography{FG2025}

\begin{thebibliography}{10}\itemsep=-1pt

\bibitem{albanie2021bbc}
S.~Albanie, G.~Varol, L.~Momeni, H.~Bull, T.~Afouras, H.~Chowdhury, N.~Fox, B.~Woll, R.~Cooper, A.~McParland, et~al.
\newblock {BBC-Oxford} {British} {sign} {language} {dataset}.
\newblock {\em arXiv preprint arXiv:2111.03635}, 2021.

\bibitem{bain2022whisperx}
M.~Bain, J.~Huh, T.~Han, and A.~Zisserman.
\newblock {WhisperX}: time-accurate speech transcription of long-form audio.
\newblock {\em INTERSPEECH 2023}, 2023.

\bibitem{bajtin1952problema}
M.~Bajt{\'\i}n.
\newblock {\em Estética de la creación verbal}, chapter El problema de los g{\'e}neros discursivos, pages 248--293.
\newblock México: Siglo XXI, 1982.

\bibitem{bragg_sign_2019}
D.~Bragg, O.~Koller, M.~Bellard, L.~Berke, P.~Boudreault, A.~Braffort, N.~Caselli, M.~Huenerfauth, H.~Kacorri, T.~Verhoef, C.~Vogler, and M.~Ringel~Morris.
\newblock Sign language recognition, generation, and translation: An interdisciplinary perspective.
\newblock In {\em Proceedings of the 21st International ACM SIGACCESS Conference on Computers and Accessibility}, ASSETS '19, page 16–31, New York, NY, USA, 2019. Association for Computing Machinery.

\bibitem{brentari1998prosodic}
D.~Brentari.
\newblock {\em {A} {Prosodic} {Model} {of} {Sign} {Language} {Phonology}}.
\newblock A Bradford book. MIT Press, 1998.

\bibitem{Bull_2021_ICCV}
H.~Bull, T.~Afouras, G.~Varol, S.~Albanie, L.~Momeni, and A.~Zisserman.
\newblock Aligning subtitles in sign language videos.
\newblock In {\em Proceedings of the IEEE/CVF International Conference on Computer Vision (ICCV)}, pages 11552--11561, October 2021.

\bibitem{bartoli_automatic_2020}
H.~Bull, M.~Gouiffès, and A.~Braffort.
\newblock Automatic {Segmentation} of {Sign} {Language} into {Subtitle}-{Units}.
\newblock In {\em Computer {Vision} – {ECCV} 2020 {Workshops}}, volume 12536, pages 186--198. Springer International Publishing, Cham, 2020.
\newblock Series Title: Lecture Notes in Computer Science.

\bibitem{camgoz2018neural}
N.~C. Camgöz, S.~Hadfield, O.~Koller, H.~Ney, and R.~Bowden.
\newblock Neural sign language translation.
\newblock In {\em 2018 IEEE/CVF Conference on Computer Vision and Pattern Recognition (CVPR)}, pages 7784--7793, 2018.

\bibitem{camgoz2020sign}
N.~C. Camgöz, O.~Koller, S.~Hadfield, and R.~Bowden.
\newblock Sign language transformers: Joint end-to-end sign language recognition and translation.
\newblock In {\em 2020 IEEE/CVF Conference on Computer Vision and Pattern Recognition (CVPR)}, pages 10020--10030, 2020.

\bibitem{carreira2016human}
J.~Carreira and A.~Zisserman.
\newblock {Quo} {Vadis}, {Action} {Recognition?} {A} new model and the {Kinetics} {Dataset}.
\newblock In {\em 2017 IEEE Conference on Computer Vision and Pattern Recognition (CVPR)}, pages 4724--4733, 2017.

\bibitem{ciapuscio1994tipos}
G.~E. Ciapuscio.
\newblock {\em Tipos textuales}.
\newblock Universidad de Buenos Aires, Argentina, 1994.

\bibitem{dal2022lsa}
P.~Dal~Bianco, G.~R{\'i}os, F.~Ronchetti, F.~Quiroga, O.~Stanchi, W.~Hasperu{\'e}, and A.~Rosete.
\newblock {LSA-T:} the first continuous argentinian sign language dataset for sign language translation.
\newblock In {\em Advances in Artificial Intelligence -- IBERAMIA 2022}, pages 293--304, Cham, 2022. Springer International Publishing.

\bibitem{de_coster_machine_2023}
M.~De~Coster, D.~Shterionov, M.~Van~Herreweghe, and J.~Dambre.
\newblock Machine translation from signed to spoken languages: state of the art and challenges.
\newblock {\em Universal Access in the Information Society}, 23(3):1305--1331, Aug. 2024.

\bibitem{de-coster-etal-2020-sign}
M.~De~Coster, M.~Van~Herreweghe, and J.~Dambre.
\newblock Sign language recognition with transformer networks.
\newblock In {\em Proceedings of the Twelfth Language Resources and Evaluation Conference}, pages 6018--6024. European Language Resources Association, 2020.

\bibitem{De_Coster_2021_CVPR}
M.~De~Coster, M.~Van~Herreweghe, and J.~Dambre.
\newblock Isolated sign recognition from {RGB} video using pose flow and self-attention.
\newblock In {\em 2021 IEEE/CVF Conference on Computer Vision and Pattern Recognition Workshops (CVPRW)}, pages 3436--3445, 2021.

\bibitem{de2015traduccao}
R.~M. de~Quadros and R.~R. Segala.
\newblock Tradu{\c{c}}{\~a}o intermodal, intersemi{\'o}tica e interlingu{\'\i}stica de textos escritos em portugu{\^e}s para a {Libras} oral.
\newblock {\em Cadernos de tradu{\c{c}}{\~a}o}, (2):354--386, 2015.

\bibitem{Duarte_CVPR2021}
A.~Duarte, S.~Palaskar, L.~Ventura, D.~Ghadiyaram, K.~DeHaan, F.~Metze, J.~Torres, and X.~Giro-i Nieto.
\newblock How2sign: A large-scale multimodal dataset for continuous american sign language.
\newblock In {\em 2021 IEEE/CVF Conference on Computer Vision and Pattern Recognition (CVPR)}, pages 2734--2743, 2021.

\bibitem{geraci2009epenthesis}
C.~Geraci.
\newblock Epenthesis in {Italian} {Sign} {Language}.
\newblock {\em Sign Language \& Linguistics}, 12(1):3--51, 2009.

\bibitem{jiang2021skeleton}
S.~Jiang, B.~Sun, L.~Wang, Y.~Bai, K.~Li, and Y.~Fu.
\newblock Skeleton aware multi-modal sign language recognition.
\newblock In {\em 2021 IEEE/CVF Conference on Computer Vision and Pattern Recognition Workshops (CVPRW)}, pages 3408--3418, 2021.

\bibitem{jm3}
D.~Jurafsky and J.~H. Martin.
\newblock {\em Speech and Language Processing: An Introduction to Natural Language Processing, Computational Linguistics, and Speech Recognition with Language Models}.
\newblock 3rd edition, 2024.
\newblock Online manuscript released August 20, 2024.

\bibitem{ko2019neural}
S.-K. Ko, C.~J. Kim, H.~Jung, and C.~Cho.
\newblock Neural sign language translation based on human keypoint estimation.
\newblock {\em Applied sciences}, 9(13):2683, 2019.

\bibitem{koller2015continuous}
O.~Koller, J.~Forster, and H.~Ney.
\newblock Continuous sign language recognition: towards large vocabulary statistical recognition systems handling multiple signers.
\newblock {\em Computer Vision and Image Understanding}, 141:108--125, 2015.

\bibitem{li2020tspnet}
D.~Li, C.~Xu, X.~Yu, K.~Zhang, B.~Swift, H.~Suominen, and H.~Li.
\newblock {TSPNet}: hierarchical feature learning via temporal semantic pyramid for sign language translation.
\newblock {\em Advances in Neural Information Processing Systems}, 33:12034--12045, 2020.

\bibitem{li2020transferring}
D.~Li, X.~Yu, C.~Xu, L.~Petersson, and H.~Li.
\newblock Transferring cross-domain knowledge for video sign language recognition.
\newblock In {\em 2020 IEEE/CVF Conference on Computer Vision and Pattern Recognition (CVPR)}, pages 6204--6213, 2020.

\bibitem{liddell1984think}
S.~K. Liddell.
\newblock Think and believe: sequentiality in {American} {Sign} {Language}.
\newblock {\em Language}, 60(2):372--399, 1984.

\bibitem{lin2004automatic}
C.-Y. Lin and F.~J. Och.
\newblock Automatic evaluation of machine translation quality using longest common subsequence and skip-bigram statistics.
\newblock In {\em Proceedings of the 42nd annual meeting of the association for computational linguistics (ACL-04)}, pages 605--612, 2004.

\bibitem{lin-etal-2023-gloss}
K.~Lin, X.~Wang, L.~Zhu, K.~Sun, B.~Zhang, and Y.~Yang.
\newblock Gloss-free end-to-end sign language translation.
\newblock In {\em Proceedings of the 61st Annual Meeting of the Association for Computational Linguistics (Volume 1: Long Papers)}, pages 12904--12916, Toronto, Canada, July 2023. Association for Computational Linguistics.

\bibitem{moryossef-etal-2023-linguistically}
A.~Moryossef, Z.~Jiang, M.~M{\"u}ller, S.~Ebling, and Y.~Goldberg.
\newblock Linguistically motivated sign language segmentation.
\newblock In H.~Bouamor, J.~Pino, and K.~Bali, editors, {\em Findings of the Association for Computational Linguistics: EMNLP 2023}, pages 12703--12724, Singapore, Dec. 2023. Association for Computational Linguistics.

\bibitem{moryossef2020real}
A.~Moryossef, I.~Tsochantaridis, R.~Aharoni, S.~Ebling, and S.~Narayanan.
\newblock Real-time sign language detection using human pose estimation.
\newblock In A.~Bartoli and A.~Fusiello, editors, {\em Computer Vision~--~ECCV 2020 Workshops}, pages 237--248, Cham, 2020. Springer International Publishing.

\bibitem{papineni2002bleu}
K.~Papineni, S.~Roukos, T.~Ward, and W.-J. Zhu.
\newblock {BLEU}: a method for automatic evaluation of machine translation.
\newblock In {\em Proceedings of the 40th annual meeting of the Association for Computational Linguistics}, pages 311--318, 2002.

\bibitem{post_2018_call}
M.~Post.
\newblock A call for clarity in reporting {BLEU} scores.
\newblock In {\em Proceedings of the Third Conference on Machine Translation: Research Papers}, pages 186--191, Brussels, Belgium, Oct. 2018. Association for Computational Linguistics.

\bibitem{Renz2021signsegmentation_a}
K.~Renz, N.~C. Stache, S.~Albanie, and G.~Varol.
\newblock Sign language segmentation with temporal convolutional networks.
\newblock In {\em ICASSP 2021~-~2021 IEEE International Conference on Acoustics, Speech and Signal Processing (ICASSP)}, pages 2135--2139, 2021.

\bibitem{shen2024auslan}
X.~Shen, S.~Yuan, H.~Sheng, H.~Du, and X.~Yu.
\newblock Auslan-{Daily}: Australian sign language translation for daily communication and news.
\newblock {\em Advances in Neural Information Processing Systems}, 36, 2024.

\bibitem{shi2021fingerspelling}
B.~Shi, D.~Brentari, G.~Shakhnarovich, and K.~Livescu.
\newblock Fingerspelling detection in {American} {Sign} {Language}.
\newblock In {\em 2021 IEEE/CVF Conference on Computer Vision and Pattern Recognition (CVPR)}, pages 4164--4173, 2021.

\bibitem{shi2022open}
B.~Shi, D.~Brentari, G.~Shakhnarovich, and K.~Livescu.
\newblock Open-domain sign language translation learned from online video.
\newblock In {\em Proceedings of the 2022 Conference on Empirical Methods in Natural Language Processing}, pages 6365--6379, Abu Dhabi, United Arab Emirates, Dec. 2022. Association for Computational Linguistics.

\bibitem{stassi2022lsu}
A.~E. Stassi, M.~Tancredi, R.~Aguirre, A.~Gómez, B.~Carballido, A.~Méndez, S.~Beheregaray, A.~Fojo, V.~Koleszar, and G.~Randall.
\newblock {LSU-DS}: an uruguayan sign language public dataset for automatic recognition.
\newblock In {\em Proceedings of the 11th International Conference on Pattern Recognition Applications and Methods (ICPRAM)}, pages 697--705. INSTICC, SciTePress, 2022.

\bibitem{tarres2023sign}
L.~Tarrés, G.~I. Gállego, A.~Duarte, J.~Torres, and X.~Giró-i Nieto.
\newblock Sign language translation from instructional videos.
\newblock In {\em 2023 IEEE/CVF Conference on Computer Vision and Pattern Recognition Workshops (CVPRW)}, pages 5625--5635, 2023.

\bibitem{varol_read_2021}
G.~Varol, L.~Momeni, S.~Albanie, T.~Afouras, and A.~Zisserman.
\newblock Read and attend: temporal localisation in sign language videos.
\newblock In {\em 2021 IEEE/CVF Conference on Computer Vision and Pattern Recognition (CVPR)}, pages 16852--16861, 2021.

\bibitem{voskou2021stochastic}
A.~Voskou, K.~P. Panousis, D.~Kosmopoulos, D.~N. Metaxas, and S.~Chatzis.
\newblock Stochastic transformer networks with linear competing units: application to end-to-end {SL} translation.
\newblock In {\em 2021 IEEE/CVF International Conference on Computer Vision (ICCV)}, pages 11926--11935, 2021.

\bibitem{wong2022hierarchical}
R.~Wong, N.~C. Camg{\"o}z, and R.~Bowden.
\newblock Hierarchical {I3D} for sign spotting.
\newblock In {\em Computer Vision -- ECCV 2022 Workshops}, pages 243--255, Cham, 2023. Springer Nature Switzerland.

\bibitem{yin2023gloss}
A.~Yin, T.~Zhong, L.~Tang, W.~Jin, T.~Jin, and Z.~Zhao.
\newblock Gloss attention for gloss-free sign language translation.
\newblock In {\em 2023 IEEE/CVF Conference on Computer Vision and Pattern Recognition (CVPR)}, pages 2551--2562, 2023.

\bibitem{zhang2019bertscore}
T.~Zhang, V.~Kishore, F.~Wu, K.~Q. Weinberger, and Y.~Artzi.
\newblock {BERTScore}: Evaluating text generation with {BERT}.
\newblock In {\em International Conference on Learning Representations (ICLR)}, 2020.

\bibitem{zhou2021improving}
H.~Zhou, W.~Zhou, W.~Qi, J.~Pu, and H.~Li.
\newblock Improving sign language translation with monolingual data by sign back-translation.
\newblock In {\em 2021 IEEE/CVF Conference on Computer Vision and Pattern Recognition (CVPR)}, pages 1316--1325, 2021.

\bibitem{Zhou_Punctuation_2022}
Z.~Zhou, T.~Tan, and Y.~Qian.
\newblock Punctuation prediction for streaming on-device speech recognition.
\newblock In {\em ICASSP 2022 - 2022 IEEE International Conference on Acoustics, Speech and Signal Processing (ICASSP)}, pages 7277--7281, 2022.

\bibitem{zhu2024chinese}
Q.~Zhu, J.~Li, F.~Yuan, J.~Fan, and Q.~Gan.
\newblock A {Chinese} continuous sign language dataset based on complex environments.
\newblock {\em arXiv preprint arXiv:2409.11960}, 2024.

\end{thebibliography}

}
\end{document}